\documentclass[runningheads]{llncs}

\usepackage[T1]{fontenc}

\usepackage{graphicx}
\usepackage{tabularx}

\begin{document}

\title{\hspace{-5mm}MetaLogic: Robustness Evaluation of Text-to-Image Models via Logically Equivalent Prompts}
\titlerunning{MetaLogic}

\author{
Yifan Shen\orcidID{0009-0007-7559-0200} \and
Yangyang Shu\orcidID{0000-0001-6319-6037} \and
Hye-young Paik\orcidID{0000-0003-4425-7388} \and
Yulei Sui\orcidID{0000-0002-9510-6574}
}

\authorrunning{Y. Shen et al.}

\institute{
  University of New South Wales, Australia\\
  \email{yifan.shen3@student.unsw.edu.au, yangyang.shu@unsw.edu.au, h.paik@unsw.edu.au, y.sui@unsw.edu.au}
}

\maketitle            

\begin{abstract}
Recent advances in text-to-image (T2I) models, especially diffusion-based architectures, have significantly improved the visual quality of generated images. However, these models continue to struggle with a critical limitation: maintaining semantic consistency when input prompts undergo minor linguistic variations. Despite being logically equivalent, such prompt pairs often yield misaligned or semantically inconsistent images, exposing a lack of robustness in reasoning and generalisation. To address this, we propose MetaLogic, a novel evaluation framework that detects T2I misalignment without relying on ground truth images. MetaLogic leverages metamorphic testing, generating image pairs from prompts that differ grammatically but are semantically identical. By directly comparing these image pairs, the framework identifies inconsistencies that signal failures in preserving the intended meaning, effectively diagnosing robustness issues in the model’s logic understanding. Unlike existing evaluation methods that compare a generated image to a single prompt, MetaLogic evaluates semantic equivalence between paired images, offering a scalable, ground-truth-free approach to identifying alignment failures. It categorises these alignment errors (e.g., entity omission, duplication, positional misalignment) and surfaces counterexamples that can be used for model debugging and refinement. We evaluate MetaLogic across multiple state-of-the-art T2I models and reveal consistent robustness failures across a range of logical constructs. We find that even the SOTA text-to-image models like Flux.dev and DALLE-3 demonstrate a 59\% and 71\% misalignment rate, respectively. Our results show that MetaLogic is not only efficient and scalable, but also effective in uncovering fine-grained logical inconsistencies that are overlooked by existing evaluation metrics.
\keywords{Metamorphic testing \and Image generation evaluation \and Semantic misalignment \and Robustness evaluation}
\end{abstract}

\section{Introduction}
Text-to-image (T2I) generation has made remarkable strides in recent years, largely due to innovations in generative architectures such as diffusion models (e.g., Imagen, GLIDE, DALLE-2)\cite{Saharia2022PhotorealisticTD}\cite{ramesh2022hierarchical}\cite{Nichol2021GLIDETP}. These models now produce high-quality and visually coherent outputs and have found applications across creative industries, education, healthcare, and architectural design. However, despite improvements in image fidelity, a critical limitation remains: \textbf{the semantic alignment between the generated image and the intended meaning of the input prompt.} In particular, these models often struggle to preserve logical consistency when semantically equivalent but syntactically different prompts are used.

This shortcoming reveals a fundamental issue of \textbf{robustness}, a reliable T2I model should generate semantically consistent images even when minor, logically equivalent variations are made to the input text. Yet, even state-of-the-art models frequently produce divergent outputs for prompts that differ only in phrasing but express the same logical content. For example, given the prompts “a cat next to a dog, with an apple” and “a dog next to a cat, with an apple,” some models omit entities or alter the spatial relationships, revealing poor generalisation under minimal semantic perturbations.

In parallel, logic has also been applied on the generation side. For instance, Sueyoshi et al. \cite{sueyoshi2024predicated} introduces Predicate Diffusion, which incorporates predicate logic constraints into diffusion models to improve prompt-image alignment. While complementary to our aims, such methods focus on logic-guided generation rather than evaluation.

Traditional evaluation methods, such as CLIPScore \cite{radford2021learning} or VQA-based frameworks (e.g., TIFA \cite{hu2023tifa}), rely on comparing a generated image to its original text prompt, treating the text as a proxy ground truth. These methods are often computationally expensive, susceptible to biases, and fail to adequately assess fine-grained logical consistency. Moreover, they do not scale efficiently to large datasets and are not designed to evaluate robustness under semantically equivalent prompt variations.

To address these limitations, we introduce MetaLogic, a novel framework that shifts the evaluation paradigm from text-image alignment to \textbf{text-to-image pairwise robustness}. Rather than comparing a single image to its corresponding prompt, MetaLogic's novelty lies in using pairs of logically equivalent prompts to generate two images and comparing them directly. If the model is logically consistent and robust, these image pairs should be semantically identical. Misalignments between the generated image pair, despite semantic equivalence in their prompts, serve as counterexamples that highlight the model's failure to maintain logical consistency and robustness. These inconsistencies were observed across 1600 images generated by state-of-the-art models such as DALLE-3 \cite{betker2023improving} and Flux.dev \cite{flux2024}, and they become increasingly pronounced as the logical complexity of prompts grows, posing significant challenges for users who expect consistent and reliable outputs.

Through extensive experiments, we show that MetaLogic not only detects misalignments due to robustness issues but also systematically categorises logical errors, including entity omission, duplication, and spatial misplacement, providing valuable insights for debugging and improving generative models. Unlike traditional approaches, MetaLogic enables scalable, oracle-free evaluation while emphasising the core requirement of robust semantic preservation under linguistic perturbation.

Our framework draws inspiration from predicate logic equivalence and metamorphic testing, a software testing methodology designed for systems without a reliable oracle. By using logically equivalent prompts as metamorphic relations, we circumvent the need for external ground truth and enable efficient, scalable evaluation of text-to-image robustness. These comparisons reveal fine-grained logical inconsistencies across a wide range of logic categories, including commutativity, associativity, distributivity, and DeMorgan transformations.

We summarise the main contributions of this work as follows:
\begin{itemize}
    \item Robustness-oriented evaluation via counterexample generation: We propose a novel method to evaluate the robustness of text-to-image (T2I) models by measuring the semantic consistency of image pairs generated from logically equivalent prompts. Our approach systematically identifies and categorises inconsistencies as counterexamples, providing concrete evidence of model errors and enabling deeper analysis of failure modes.
    \item Ground truth-free detection of alignment errors: Our metamorphic testing approach avoids reliance on external labels by treating one generated image as the oracle for another, enabling scalable and efficient evaluation.
    \item Comprehensive analysis of state-of-the-art models: We apply our framework to two leading T2I models, Flux.dev and DALLE-3, evaluating their robustness across five categories of logical equivalence; Associative, Commutative, DeMorgan, Distributive, and Complement. The analysis is based on three semantic criteria: object presence, spatial positioning, and entity count, providing a multifaceted view of model performance under logical perturbations.
\end{itemize}

\section{Preliminaries}
In this section, we discuss the necessary preliminary knowledge and prior work, highlight their differences from ours, and then present our approach and methodology in detail.

\noindent\textbf{Zero-shot image classifiers.}
CLIP (Contrastive Language–Image Pretraining) \cite{radford2021learning} is a widely used benchmark for text-image alignment, trained on millions of image-text pairs to learn a joint embedding space where related images and texts are nearby. CLIPScore measures alignment via cosine similarity between embeddings. As a zero-shot classifier, CLIP works without ground truth labels, offering flexibility across tasks. However, it struggles with fine-grained semantics, logical reasoning, and complex prompts \cite{ahmadi2023examination}. Consequently, in the context of this work, which focuses on detecting subtle logical inconsistencies, CLIP is used only as a coarse-grained baseline, not as a reliable measure for logic-sensitive evaluation.

\noindent\textbf{VQA-based alignment evaluation.}
The TIFA framework \cite{hu2023tifa} evaluates text-image alignment using Vision Question Answering (VQA) models. An LLM generates question-answer pairs from the text prompt, and a VQA model answers these questions based on the image; alignment is measured by comparing the answers. However, generating QA pairs and VQA answers for each image is computationally intensive and limits scalability. Similarly, Singh et al. \cite{singh2023divide} propose Decompositional-Alignment-Score, where an LLM decomposes the prompt into factual assertions evaluated by a VQA model. While improving interpretability, it faces the same scalability challenges as TIFA.

\noindent\textbf{Object-based evaluation.}
Object detection-based frameworks evaluate text-image alignment by comparing generated images against structured ground truth from standardised templates, as seen in TIAM \cite{grimal2024tiam} and GenEval \cite{ghosh2024geneval}. While more computationally efficient than VQA-based methods, their reliance on templates and detectable entities limits flexibility with open-ended prompts. GenEval further categorises prompts into tasks (e.g., object recognition, counting, colour, spatial relations), enabling fine-grained diagnostics and identifying specific model failures for targeted improvements.

\noindent\textbf{Logic-based evaluation}
Recent work has integrated logical reasoning into AI evaluation. LogicAsker \cite{wan-etal-2024-logicasker} tests LLMs using logic-based challenges (e.g., propositional logic, quantifiers, conditionals), with correctness judged against ground truth answers. In contrast, our MetaLogic framework removes the need for ground truth by applying metamorphic testing: it generates logically equivalent prompts in different forms, and misalignments between the resulting images indicate failures in logical consistency. This approach is well-suited for T2I models, where ground truth is often missing. Unlike LogicAsker’s textual focus, our MetaLogic enables robust, oracle-free evaluation of logical consistency. 

\noindent\textbf{Metamorphic testing}
MetaLogic is grounded in predicate-logic-based metamorphic testing, a technique used when a reliable output oracle is unavailable~\cite{8573811}\cite{chen2020metamorphictestingnewapproach}\cite{10.1145/3143561}. Instead of verifying individual outputs, it evaluates the consistency between outputs by leveraging known input-output relationships, called metamorphic relations. For example, since $\sin(x) = \sin(\pi - x)$, a program claiming to compute $\sin(x)$ can be tested by comparing these outputs without needing exact ground truth. Similarly, MetaLogic tests T2I models by generating logically equivalent prompts and checking for alignment in the resulting images, revealing failures when consistency breaks.

\noindent\textbf{Reasoning capabilities in AI models.}
A fundamental component of understanding logic is a high level of reasoning capabilities \cite{bronkhorst2020logical}. High-level reasoning is a core step in the gradual evolution of deep learning models, evident from the release of modern reasoning models, a progression from early large language models \cite{huang2022towards}. A recent study by \cite{mirzadeh2024gsm} puts into question the robustness of mathematical reasoning in modern large language models. They observed that by introducing additional clauses, irrelevant to the process of reaching the final answer, a significant performance drop occurred across all state-of-the-art models. The observed fragility of current LLMs makes one question their relevance to other models in close proximity, such as text-to-image.

\begin{figure}[t]
    \centering
    \includegraphics[width=\textwidth]{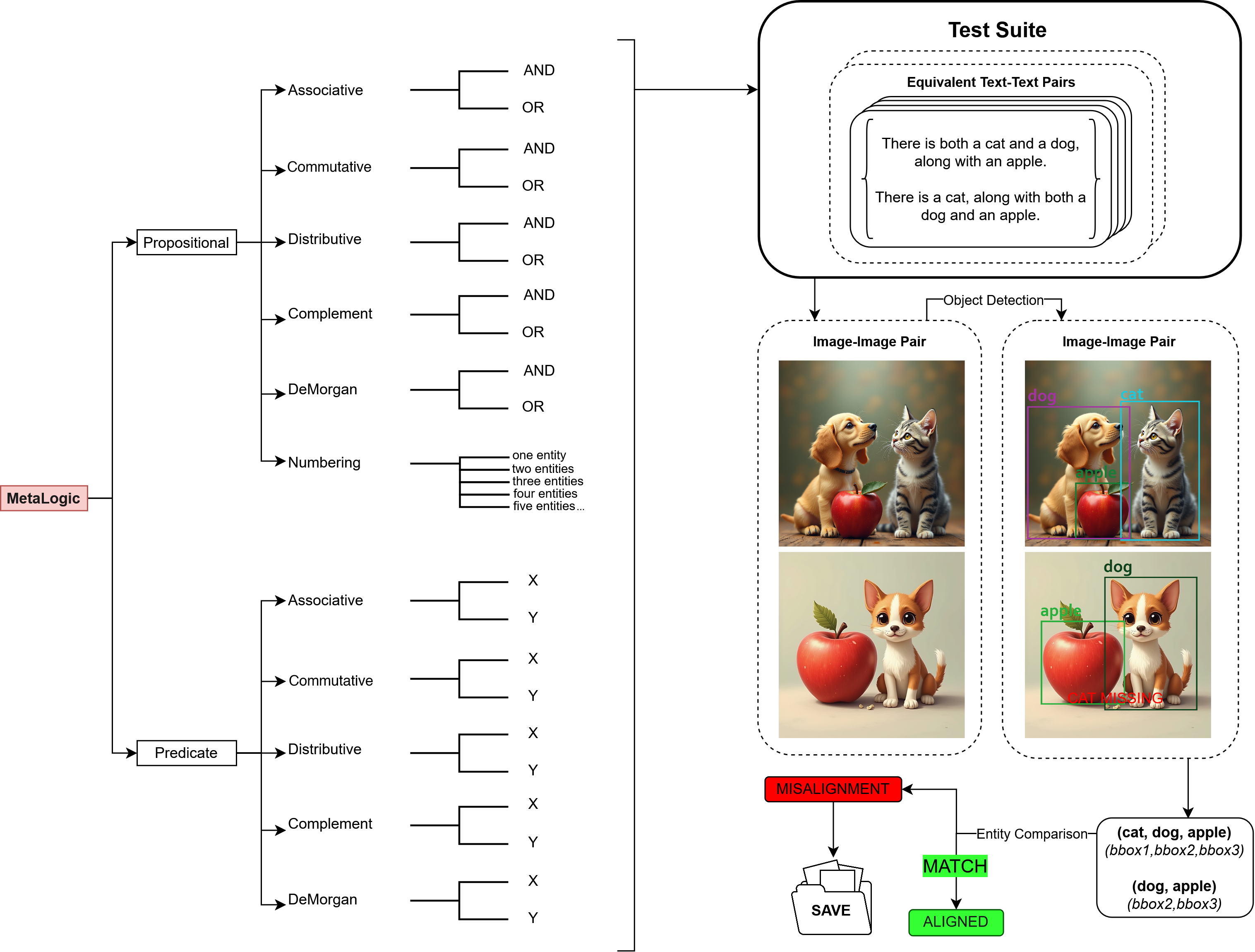}
    \caption{Outline of MetaLogic framework which is split into two main sections. The left tree diagram shows the breakdown of all logic prompt categories and how they relate to each other. The right side outlines the misalignment detection process where we compare two images, generated from equivalent text pairs. Any detected inconsistencies are logged and saved as a counterexample. } 
    \label{fig:framework}
    \vspace{-5mm}
\end{figure}

\vspace{-2mm}
\section{Our Approach}
\vspace{-2mm}
We introduce a new framework in Figure~\ref{fig:framework} to quickly and precisely identify text-image misalignment due to model robustness issues and procure effective counterexamples for model improvement. With speed and efficiency at the forefront, we present a methodology based on comparing image-image pairs instead of text-image pairs seen in existing methods. To facilitate the image-image comparison, we employ logic equivalence laws in each text-text pair. Hence, the text pairs for each image pair would be grammatically different but semantically equivalent. A fully robust T2I model would perfectly translate the semantic equivalence in each text pair into semantically identical image pairs. Therefore, inconsistencies between image pairs infer the existence of a text-image misalignment in either text-image pair. 

First, we determine the compatible logic equivalence laws to apply to our set of prompt pairs. We then generate multiple prompt pairs based on the previously determined equivalence laws. Lastly, we generate images from each logically equivalent prompt pair and compare the image-image pairs. Inconsistencies between image-image pairs are noted as text-image misalignment and taken as counterexample data for future model improvements (See Figure \ref{fig:framework}).

\vspace{-2mm}
\subsection{Logic equivalence laws}
\vspace{-2mm}
The metamorphic relationship found in trigonometry can be applied to the realm of natural language in the form of logically equivalent sentence pairs. Logical equivalence can be defined as when two statements have the same semantic meaning but are written in different forms.

\begin{table}[ht]
    \centering
    \renewcommand{\arraystretch}{0.95}
    \setlength{\tabcolsep}{6pt}
    \begin{tabularx}{\linewidth}{
        p{1.5cm} 
        >{\centering\arraybackslash}p{3cm} 
        >{\centering\arraybackslash}p{2.3cm} 
        p{4cm}   
    }
    \hline
    \textbf{Law} & \textbf{Logical Equivalence} & \textbf{Prompt Transformation} & \textbf{Prompt Example} \\
    \hline
    Commutative & $P \wedge Q \Leftrightarrow Q \wedge P$ & Reordering entities & There is a cat and a dog $\Leftrightarrow$ There is a dog and a cat \\
    & $P \vee Q \Leftrightarrow Q \vee P$ & Reordering entities & There is a cat or a dog $\Leftrightarrow$ There is a dog or a cat \\ \hline
    Associative & $(P \wedge Q) \wedge R \Leftrightarrow P \wedge (Q \wedge R)$ & Shifting semantic pairing & There is both a cat and a dog, along with an apple $\Leftrightarrow$ There is a cat, along with both a dog and an apple \\
    & $(P \vee Q) \vee R \Leftrightarrow P \vee (Q \vee R)$ & Shifting semantic pairing & There is either a cat or a dog, otherwise there is an apple $\Leftrightarrow$ There is a cat, otherwise there is a dog or an apple \\ \hline
    Distributive & $P \wedge (Q \vee R) \Leftrightarrow (P \wedge Q) \vee (P \wedge R)$ & Sharing entity across group & There is a cat with either a dog or an apple $\Leftrightarrow$ There is a cat with a dog or a cat with an apple \\
    & $P \vee (Q \wedge R) \Leftrightarrow (P \vee Q) \wedge (P \vee R)$ & Sharing entity across group & There is a cat or there is a dog and an apple $\Leftrightarrow$ There is a cat or a dog or there is a cat or an apple \\ \hline
    Complement & $\neg(\neg P) \Leftrightarrow P$ & Double negatives & It is not true that there is not a cat $\Leftrightarrow$ There is a cat \\ \hline
    DeMorgan & $\neg(P \wedge Q) \Leftrightarrow \neg P \vee \neg Q$ & Sharing negation and sign flip & There isn't a cat and a dog $\Leftrightarrow$ There isn't a cat or there isn't a dog \\
    DeMorgan & $\neg(P \vee Q) \Leftrightarrow \neg P \wedge \neg Q$ & Sharing negation and sign flip & There isn't a cat or a dog $\Leftrightarrow$ There isn't a cat and there isn't a dog \\ 
    \hline
    \end{tabularx}
    \caption{Logic equivalence law categories used with their respective mathematical notation, prompt transformation condition and natural language example.}
    \label{tab:categories}
    \vspace{-7mm}
\end{table}

\noindent{\textbf{Propositional logic law selection}.}
The framework required a large suite of test cases to analyse in order to test a given T2I model. Luckily, logical equivalence is categorised into discrete laws that behave differently from each other. We list all the logic law categories considered in Table \ref{tab:categories}. Early on, a problem with the initial method theory was found that involved disjunctive (OR) statements. An OR statement allowed for either subject to correctly exist in a generated image and be considered correct. However, with our image comparison framework, if even two correct outcomes were to exist for each prompt in a pair, a hypothetically infallible model would still show ``errors'' half of the time. Multiple expected outcomes break the underlying principle of metamorphic testing, and forcibly implementing them would be outside our current scope. We came up with a workaround to overcome this problem, outlined in the following section.
Additionally, the Idempotent law was removed due to oversimplification and lack of characteristics to scale efficiently. 
The final chosen logic equivalence law categories were \textit{Commutative, Associative, Distributive, Complement, and DeMorgan}. (See Table \ref{tab:categories})

\noindent\textbf{The ``disjunctive'' problem}
As stated previously, a significant challenge in applying metamorphic testing to OR statements emerged early in our methodology development. Disjunctive (OR) statements inherently allow for multiple valid interpretations if a prompt states ``There is a cat or a dog,'' both an image containing only a cat or an image containing only a dog would be considered correct. Multiple valid outcomes fundamentally conflicts with the metamorphic testing principle, which requires consistent outputs for equivalent inputs. To address this limitation while still including disjunctive logic in our evaluation, we developed a workaround: embedding conjunctive (AND) statements within disjunctive (OR) statements. For example, instead of simply testing "There is a cat or a dog" against "There is a dog or a cat," we test "There is [(a cat and a dog)] or [(a dog and a cat)]" against "There is [(a dog and a cat)] or [(a cat and a dog)]" (See Table \ref{tab:templates_base}). This approach ensures that both sides of the OR statement contain identical conjunctive expressions that differ only in grammatical structure but maintain semantic equivalence. By structuring OR statements this way, we preserve the metamorphic relationship while still evaluating the model's handling of disjunctive logic. This technique allowed us to expand our testing scope to include all major logical operations while maintaining the validity of our comparison methodology.

\begin{table}[ht!]
  \centering
  \begin{tabularx}{\textwidth}{lX}
    \hline
    \textbf{Law} & \textbf{Prompt pair template (e = entity)} \\
    \hline
    {Commutative (AND)} & {There is \textbf{(e1)} and \textbf{(e2)}} $\Leftrightarrow$ {There is \textbf{(e2)} and \textbf{(e1)}}        \\
    {Commutative (OR)} & {There is \textbf{[(e1) and (e2)]} or \textbf{[(e2) and (e1)]} $\Leftrightarrow$ {There is \textbf{[(e2) and (e1)]} or \textbf{[(e1) and (e2)]}.}} \\\hline
    {Associative (AND)} & {There is both \textbf{[(e1) and (e2)]}, along with \textbf{(e3)} $\Leftrightarrow$ {There is \textbf{(e1)}, along with both \textbf{[(e2) and (e3)]}} }         \\
    {Associative (OR)} & {There is either \textbf{[(e1) and (e2) and (e3)]} or \textbf{[(e1) and (e3) and (e2)]} ,otherwise there is \textbf{[(e2) and (e1) and (e3)]} $\Leftrightarrow$}\\
                & {There is \textbf{[(e1) and (e2) and (e3)]}, otherwise there is \textbf{[(e1) and (e3) and (e2)]}} or {\textbf{[(e2) and (e1) and (e3)]}}\\\hline
    {Distributive (AND)} & {There is \textbf{(e1)} with either both \textbf{[(e2) and (e3)]} or both \textbf{[(e3) and (e2)]}} $\Leftrightarrow$\\
    {} & {There is \textbf{(e1)} with both \textbf{[(e2) and (e3)]} or \textbf{(e1)} with both \textbf{[(e3) and (e2)]}}\\
    {Distributive (OR)} & {There is either \textbf{[(e1), (e2) and (e3)]} or both \textbf{[(e1), (e3) and (e2)]} and \textbf{[(e2), (e3) and (e1)]} $\Leftrightarrow$} \\
    {} & {There is either \textbf{[(e1), (e2) and (e3)]} or \textbf{[(e1),(e3) and (e2)]},} \\
    {} & {and there is either \textbf{[(e1), (e2) and (e3)]} or \textbf{[(e2), (e3) and (e1)]}.}\\\hline
  \end{tabularx}
  \caption{The baseline standard prompt pair templates used in the test suite.}
  \label{tab:templates_base}
  \vspace{-3mm}
\end{table}

\begin{table*}[ht!]
  \centering
  \begin{tabularx}{\textwidth}{lX}
    \hline
    \textbf{Law} & \textbf{Prompt pair template (e = entity)} \\
    \hline
    {Commutative (X)} & {There is \textbf{(e1) on the right} and \textbf{(e2) on the left}} $\Leftrightarrow$ {There is \textbf{(e2) on the left} and \textbf{(e1) on the right}}        \\
    {Commutative (Y)} & {There is \textbf{(e1) on the bottom} and \textbf{(e2) on top}} $\Leftrightarrow$ {There is \textbf{(e2) on top} and \textbf{(e1) on the bottom}} \\\hline
    {Associative (X)} & {There is both \textbf{[(e1) on the right and (e2) on the left]}, along with \textbf{(e3) in the middle}} $\Leftrightarrow$ \\
    {} & {{There is \textbf{(e1) on the right}, along with both \textbf{([(e2) on the left and (e3) in the middle]}} }         \\
    {Associative (Y)} & {There is both \textbf{[(e1) on the bottom and (e2) on top]}, along with \textbf{(e3) in the middle}} $\Leftrightarrow$ \\
    {} & {{There is \textbf{(e1) on the bottom}, along with both \textbf{[(e2) on top and (e3) in the middle]}} }     \\\hline
    {Distributive (X)} & {There is \textbf{(e1) on the right} with either both \textbf{[(e2) on left and (e3) in the middle]} or both } \\
    {} & {\textbf{[(e3) in the middle and (e2) on the left] $\Leftrightarrow$ There is \textbf{(e1) on the right} with both \textbf{[(e2) on the left and (e3) in the middle]}}} \\
    {} & { or \textbf{(e1) on the right} with both \textbf{[(e3) in the middle and (e2) on the left]}}\\
    {Distributive (Y)} & {There is \textbf{(e1) on the bottom} with either both \textbf{[(e2) on top and (e3) in the middle]} or both } \\
    {} & {\textbf{[(e3) in the middle and (e2) on top] $\Leftrightarrow$ There is \textbf{(e1) on the bottom} with both \textbf{[(e2) on top and (e3) in the middle]}}} \\
    {} & { or \textbf{(e1) on the bottom} with both \textbf{[(e3) in the middle and (e2) on top]}}\\\hline
  \end{tabularx}
  \caption{The baseline positional prompt pair templates used in the test suite.}
  \label{tab:templates_position}
    \vspace{-0mm}

\end{table*}

\begin{table*}[ht!]
  \centering
  \begin{tabularx}{\textwidth}{lX}
    \hline
    \textbf{Law} & \textbf{Prompt pair template (e = entity)} \\
    \hline
    {Complement (AND)} & {There is \textbf{(e1)} and \textbf{(e2)}.} $\Leftrightarrow$ {It is not the case that there is not \textbf{(e1)} and \textbf{(e2)}.}        \\
    {Complement (OR)} & {There is \textbf{(e1) and (e2)]} or \textbf{[(e2)} and (e1)] $\Leftrightarrow$ It is not the case that there is not \textbf{[(e1) and (e2)]} or \textbf{[(e2) and (e1)]}.} \\
    {Complement (X)} & {There is \textbf{[(e1) on the right]} and \textbf{[(e2) on the left]} $\Leftrightarrow$ {It is not the case that there is not \textbf{[(e1) on the right]} and \textbf{[(e2) on the left]}}}       \\
    {Complement (Y)} & {There is \textbf{[(e1) on the bottom]} and \textbf{[(e2) on top]} $\Leftrightarrow$ {It is not the case that there is not \textbf{[(e1) on the bottom]} and \textbf{[(e2) on top]}}}      \\\hline
    {DeMorgan (AND)} & {It is not the case that there is \textbf{[no (e1) and no (e2)]} and \textbf{[no (e2) and no (e1)]} $\Leftrightarrow$ } \\
    {} & {There isn't \textbf{[no (e1) and no (e2)]} or there isn't \textbf{[no (e2) and no (e1)]}  } \\
    {DeMorgan (OR)} & {It is not the case that there is \textbf{[no (e1) and no (e2)]} or \textbf{[no (e2) and no (e1)]} $\Leftrightarrow$ } \\
    {} & {There isn't \textbf{[no (e1) and no (e2)]} and there isn't \textbf{[no (e2) and no (e1)]}  } \\
    {DeMorgan (X)} & {It is not the case that there is \textbf{[no (e1) on the right and no (e2) on the left]} and \textbf{[no (e2) on the left and no (e1) on the right]} $\Leftrightarrow$ }\\
    {} & {There isn't \textbf{[no (e1) on the right and no (e2) on the left]} or there isn't \textbf{[no (e2) on the left and no (e1) on the right]}  } \\
    {DeMorgan (Y)} & {It is not the case that there is \textbf{[no (e1) on the bottom and no (e2) on top]} and \textbf{[no (e2) on top and no (e1) on the bottom]} $\Leftrightarrow$ }\\
    {} & {There isn't \textbf{[no (e1) on the bottom and no (e2) on top]} or there isn't \textbf{[no (e2) on top and no (e1) on the bottom]}  } \\ \hline
  \end{tabularx}
  \caption{The baseline negation prompt pair templates used in the test suite.}
  \label{tab:templates_negation}
    \vspace{-3mm}
\end{table*}

\begin{table*}[ht!]
  \centering
  \begin{tabularx}{\textwidth}{lX}
    \hline
    \textbf{Law} & \textbf{Prompt pair template (e = entity)} \\
    \hline
    {Numbering (1)} & {There is a \textbf{(e1)} and \textit{one} \textbf{(e2)}.} $\Leftrightarrow$ {There is \textit{one} \textbf{(e1)} and a \textbf{(e2)}.}        \\
    {Numbering (2)} & {There is a \textbf{(e1)} and \textit{two} \textbf{(e2)}.} $\Leftrightarrow$ {There is \textit{two} \textbf{(e1)} and a \textbf{(e2)}.}        \\
    {Numbering (3)} & {There is a \textbf{(e1)} and \textit{three} \textbf{(e2)}.} $\Leftrightarrow$ {There is \textit{three} \textbf{(e1)} and a \textbf{(e2)}.}        \\
    {Numbering (4)} & {There is a \textbf{(e1)} and \textit{four} \textbf{(e2)}.} $\Leftrightarrow$ {There is \textit{four} \textbf{(e1)} and a \textbf{(e2)}.}        \\
    {Numbering (5)} & {There is a \textbf{(e1)} and \textit{five} \textbf{(e2)}.} $\Leftrightarrow$ {There is \textit{five} \textbf{(e1)} and a \textbf{(e2)}.}        \\
    {Numbering (6)} & {There is a \textbf{(e1)} and \textit{six} \textbf{(e2)}.} $\Leftrightarrow$ {There is \textit{six} \textbf{(e1)} and a \textbf{(e2)}.}        \\
    {Numbering (7)} & {There is a \textbf{(e1)} and \textit{seven} \textbf{(e2)}.} $\Leftrightarrow$ {There is \textit{seven} \textbf{(e1)} and a \textbf{(e2)}.}        \\
    {Numbering (8)} & {There is a \textbf{(e1)} and \textit{eight} \textbf{(e2)}.} $\Leftrightarrow$ {There is \textit{eight} \textbf{(e1)} and a \textbf{(e2)}.}        \\
    {Numbering (9)} & {There is a \textbf{(e1)} and \textit{nine} \textbf{(e2)}.} $\Leftrightarrow$ {There is \textit{nine} \textbf{(e1)} and a \textbf{(e2)}.}        \\
    {Numbering (10)} & {There is a \textbf{(e1)} and \textit{ten} \textbf{(e2)}.} $\Leftrightarrow$ {There is \textit{ten} \textbf{(e1)} and a \textbf{(e2)}.}        \\
    \hline
  \end{tabularx}
  \caption{The baseline number prompt pair templates used in the test suite.}
  \label{tab:templates_numbering}
    \vspace{-1mm}
\end{table*}

\subsubsection{Combination prompting.}
To vertically scale the three chosen logic laws, we increase the number of unique prompts without increasing length. We do this by creating a list of entities and creating unique combinations of text prompt pairs guided by each logic category.

Our implementation of combination prompting is similar to techniques such as dynamic prompting found in recent literature \cite{Mo_2024_CVPR}\cite{10531644}\cite{yang2023dynamic}. While these methods may differ in terminology, they share a common objective: to expand the test space by systematically generating prompt variations from a common baseline.

Since we are testing how well any given T2I model understands logical nuances in natural language, the scope of entities is not of great concern. We seek failure by misunderstanding logic, not by mistaking the likeness of an exotic object or animal. Hence, we refer to the common entities that appear in the widely used MS COCO data set \cite{lin2014microsoft}. The entities chosen were a cat, a dog, an apple, a banana and a cow. As MS COCO is commonly used as training data and an evaluation tool, we determined that sharing similar object prompting would benefit our methodology.

\subsubsection{Predicate logic.}
Due to a number of logic equivalence laws being incompatible with metamorphic testing, further variability was required to extend the scope that our framework tested on T2I models. Up to this point, only propositional logic was used, so in an effort to increase variety, positional predicates were introduced. This not only increased our testing scope but also added positional understanding as an additional criteria in which we assess T2I models.

\subsubsection{Numbering prompts.}
To evaluate the models' ability to handle multiple instances of the same entity, we developed ``numbering prompts'' using a core set of four entities: cat, dog, apple, and banana. These prompts explicitly specified multiple occurrences of entities (e.g., ``There are two cats and a dog'' $\Leftrightarrow$ ``There is a dog and two cats''). We based these prompts on conjunctive commutative logic due to its simplicity, ensuring that the equivalence relationship remained straightforward and consistent. This approach allowed us to test whether T2I models could maintain proper entity counts while handling basic logical relationships, providing insights into how these models process quantitative information alongside qualitative descriptions. For each entity, we created prompt pairs specifying one through five instances, allowing us to track alignment rates as entity counts increased.

\begin{figure*}[ht]
    \centering
    \includegraphics[width=0.65\textwidth]{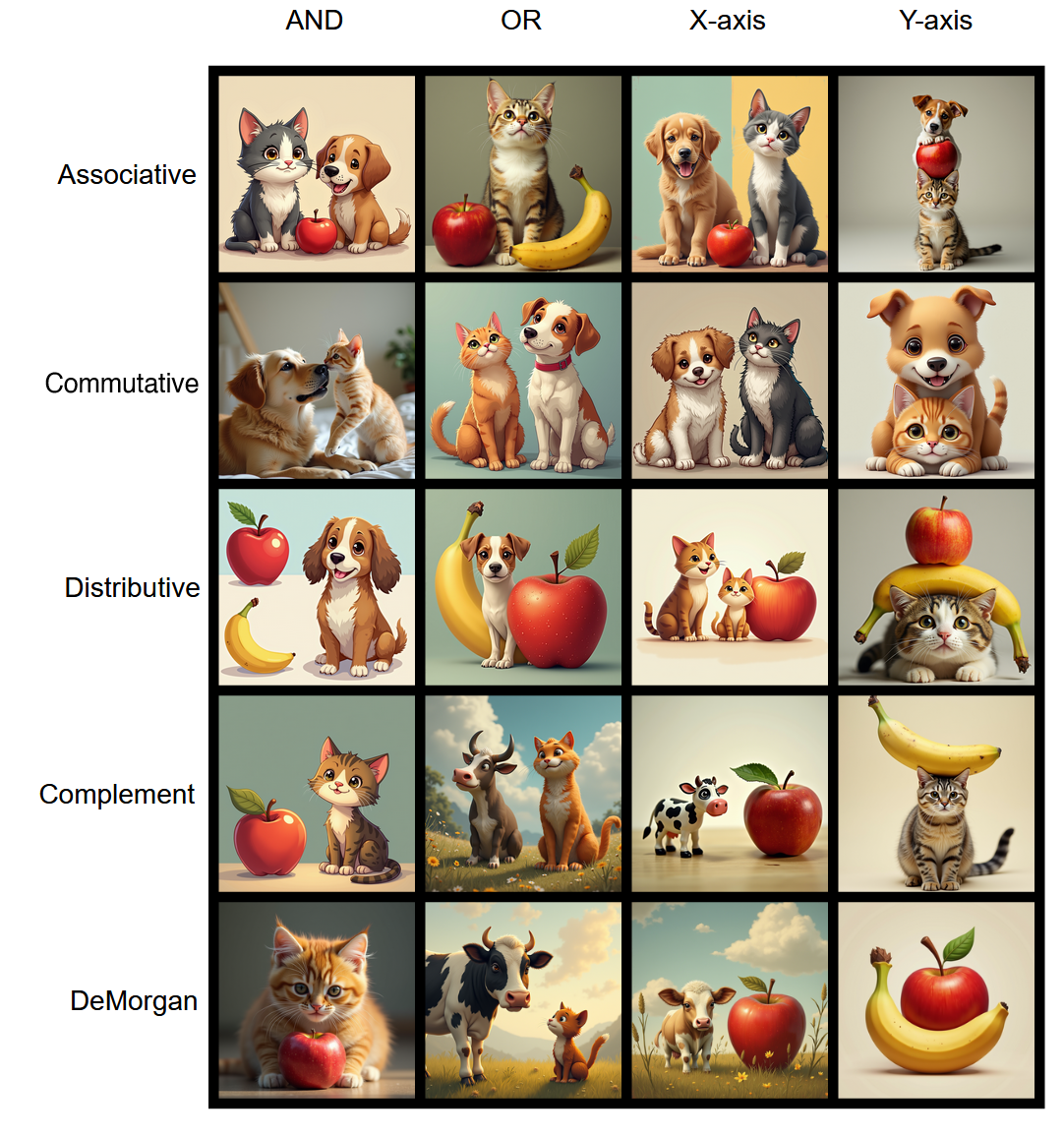}
    \vspace{-5mm}
    \caption{Illustrative example of variations in logic equivalence law categories.}
    \label{fig:image_category_examples}
    \vspace{-3mm}
\end{figure*}

\subsection{Prompt pair generation}
The generation of logic prompt pairs were divided into the five chosen equivalence laws, Associative, Commutative, Complement, Distributive and DeMorgan. These sections were then further subdivided into conjunctive (AND), disjunctive (OR), horizontal (X-axis) and vertical (Y-axis) positioning, resulting in 20 categories (See Figure \ref{fig:image_category_examples}). The numbering prompts also span across four entities and 10 numbers, totalling 40 categories. Overall, we have 60 categories, hence 60 equivalent prompt pair templates to evaluate any text-to-image model (See Tables \ref{tab:templates_base}--\ref{tab:templates_numbering})

\noindent\textbf{Object-focused prompts.}
Object-focused prompts otherwise known as prompt pairs with unspecified positioning acts as the standard and baseline for more complex prompts. The five categories of logic equivalence laws were converted into prompt templates that retain the same semantic equivalence but in the form of natural language. Each logic category is split into AND/OR sections which each hold a different template, totalling 16 templates for object-focused prompting (See Table \ref{tab:templates_base}). A Python script was written for each equivalence law to generate all possible combinations of prompt pairs using the chosen list of entities.

\noindent\textbf{Position-focused prompts.}
As stated previously, to increase the scope of testing, we incorporated positional information into the prompt pairs. By leveraging the bounding box output of object-detection models, we are also able to compare the relative position of any given entity in an image (See Table \ref{tab:templates_position}).

\noindent\textbf{Numbering-focused prompts.}
To test the numbering capabilities of text-to-image models, we modified the simplest prompt template, Commutative (AND), seen in Table \ref{tab:templates_base} and introduced a number from 1--10 to one of the two entities present in the prompt. This allows us to isolate the models capabilities in both ascending entity counts as well as potential differences between types of entities. See Table \ref{tab:templates_numbering} for all prompt templates used in the numbering category.

\vspace{-2mm}
\subsection{Image pair comparison}
\vspace{-2mm}
This section will outline the methodology taken to evaluate the image pair generated from the test suite. Our approach leverages the principle that semantically equivalent prompts produce semantically equivalent images, thereby allowing each generated image pair to serve as validation for each other. Therefore, the choice of image comparison to evaluate semantic equivalence is crucial in determining the overall efficacy of the framework.

\noindent\textbf{Object-detection vs VQA vs VLM.}
As outlined in Section 1, the current literature highlights two dominant approaches for image evaluation: methods based on Visual Question Answering (VQA) models and those using Object Detection models. A third alternative, Vision-Language Models (VLMs), was also considered. However, our approach is orthogonal to object detection, and any detection method can be selected as the method for semantic image comparison due to its ability to produce consistent and quantifiable output.

Unlike VQA and VLMs, which generate rich but  qualitative natural language descriptions, object detection yields structured outputs in the form of labelled objects and bounding boxes. This structured representation enables reliable and repeatable comparisons between images, which is important for quickly evaluating semantic alignment. The variability and subjectivity in language-based inferences from VQA and VLMs make them less suitable for tasks that require consistent, measurable comparison \cite{Yuan_2021_CVPR}. Hence, object detection provides a more efficient, robust and interpretable way for our semantic image comparison.

\noindent\textbf{Position-less image pair assessment.}
From position-unspecified prompts, we define semantic equivalence in an image pair as two images that contain the same entity labels, as well as having an equal number of total entities. Using Florence-2's object detection output, we create an entity matrix containing all labels and respective instances present within both images in the pair. The two corresponding matrices are compared with each other, noting any detected inconsistencies to represent the presence of text-image misalignment in either pair of text-image groups.

\noindent\textbf{Position pair assessment.}
Following on from comparing entities, we evaluate the spatial consistency of matching entities between image pairs. By taking the bounding box outputs provided by Florence-2, we calculate the centroid for every entity and take it as the reference point for determining relative positions from one another. These calculations are done on the x or y axis, depending on whether the initial prompt pair specified vertical or horizontal positional predicates.

\noindent\textbf{Object-detection accuracy}
Since our evaluation relies on object detection, we examined the reliability of Florence-2 by randomly sampling misaligned pairs flagged by the framework. To validate accuracy, we manually inspected random samples of 10 pairs of misaligned images in all categories. Florence-2 demonstrated around a ~90\% success rate in flagging genuine misaligned image pairs. Although not a full benchmark, these checks suggest that Florence-2 provides sufficiently reliable outcomes for our diagnostic purposes.

\vspace{-2mm}
\section{Experiments}
\vspace{-2mm}
In this section, we will discuss our experimental setup, and showcase the effectiveness of our approach with experimental results. The goal of these experiments is to evaluate the effectiveness of our methodology at exposing flaws in semantic robustness in SOTA models like DALLE-3 and Flux.dev. 

\vspace{-4mm}
\subsection{Experimental setup}
\vspace{-2mm}
To evaluate the effectiveness of our proposed framework, MetaLogic, we apply it to two state-of-the-art text-to-image (T2I) models representing both open-source and proprietary systems: Flux.dev \cite{flux2024} and DALLE-3 \cite{betker2023improving}, respectively.

\vspace{-2mm}
\subsubsection{Model setup}
\vspace{-2mm}

For both Flux.dev and DALLE-3, the images were generated through the API using a Python script in the VSCode IDE. DALLE-3 naturally expands on the initial prompt given as a means of adding detail and complexity to the final image. In our case, we want to avoid this situation as much as possible to avoid introducing unwanted variables that can sway our embedded semantic equivalence. We follow the DALLE-3 documentation provided by OpenAI to prevent any prompt alterations by including the phrase ``\textbf{I NEED to test how the tool works with extremely simple prompts. DO NOT add any detail, just use it AS-IS:}'' as a prefix to all input prompts. This prompt control technique has also been adopted in prior work to reduce generation bias \cite{conwell2024relations,fu2024commonsense}.

Flux, regarded as one of the most advanced open-source T2I models, is available in two variants: Schnell, optimised for fast generation, and Dev, designed for higher image quality at the cost of speed. For our experiments, we selected the Flux.dev version to avoid trade-offs in image fidelity.

We generate 1600 images across both models from the 800 logic prompt pairs in the test suite. The prompts pairs consist of five logic law categories (Associative, Commutative, Distributive, Complement, DeMorgan) across 4 modifiers (AND, OR, X, Y). Additionally, the numbering category spans across 4 entities (cat, dog, apple, banana) up to a count of 10. 

For the object detection, Florence-2, the base, non-fine-tuned version was used as it demonstrated the highest degree of accuracy at labelling entities from manual inspection of generated image pair samples \cite{xiao2024florence}. The model was loaded in Python script through the Hugging Face transformer library.

\vspace{-3mm}
\subsection{Effectiveness of MetaLogic}
\begin{figure}
    \centering
    \includegraphics[width=1\linewidth]{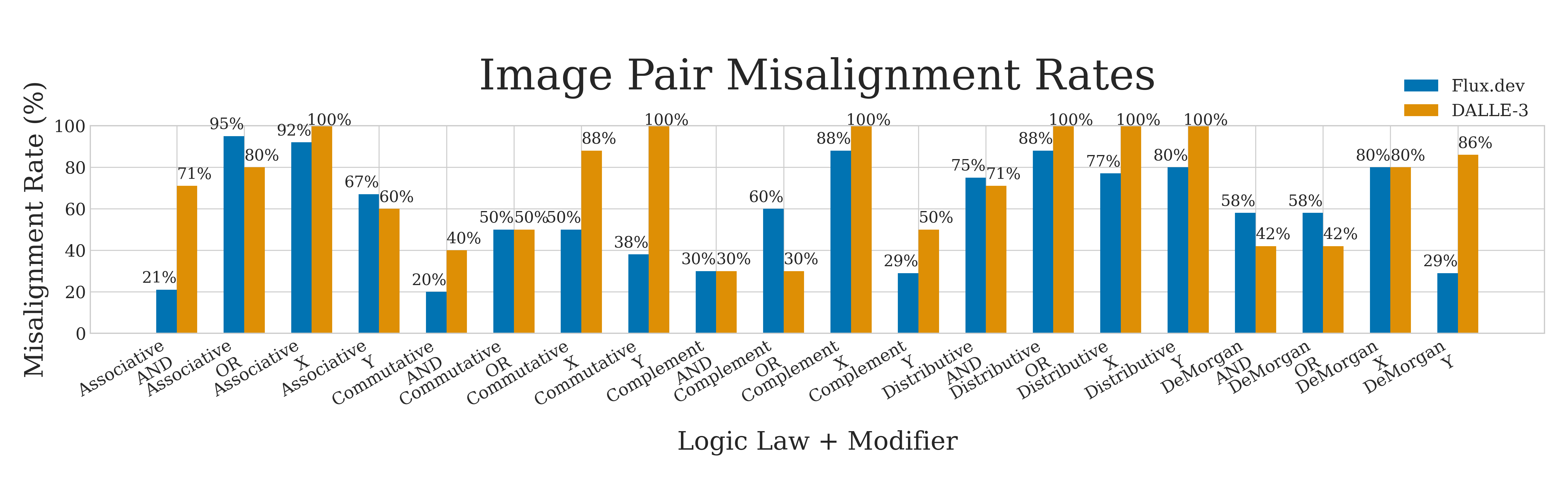}
    \vspace{-6mm}
    \caption{The misalignment rates of all 5 logic law categories (Associative, Commutative, Distributive, Complement, DeMorgan) across all 4 modifiers (AND, OR, X, Y) for both Flux.dev and DALLE-3.}
    \label{fig:mislaignment_chart}
    \vspace{-3mm}
\end{figure}

\vspace{-3mm}
Our experimental evaluation shows that MetaLogic effectively identifies logical misalignment in text-to-image generation, with detection rates varying significantly between different logical categories and their subsequent variations (See Figure \ref{fig:mislaignment_chart}).

Overall, MetaLogic detected logical inconsistencies in 65\% of generated image pairs. Performance varied substantially across logical categories, with the highest misalignment detection rates occurring in the Distributive category, followed by Associative, DeMorgan, Complement, and Commutative. This trend of increasing misalignment rates runs adjacent to the categories in order of logical complexity. This suggests that the models struggle with prompts involving complex logical structures where entities are distributed across different groups and clauses.

A clear pattern emerged when comparing AND versus OR variants across all logical categories. AND-based prompts resulted in 45.8\% misalignment rate overall, while OR-based prompts showed significantly higher inconsistency at 65.3\%. This difference suggests that the model struggles more with disjunctive logic than conjunctive logic. We likely attribute this to the grammatically complex nature of the Associative(OR) and Distributive(OR) prompts required to sustain consistent semantic equivalence (See Table \ref{tab:templates_base}). 

The addition of positional constraints had significant effects on the overall misalignment rate. Standard conjunctive (AND) prompts without positional information showed only a 21\% misalignment rate in Flux.dev, while prompts with explicit positional information (both horizontal and vertical) showed a substantial increase to 63\% on average for vertical, up to 85\% for horizontal position prompts. Surprisingly, the X modifier performed much worse than Y where you would expect them to be similar. Nevertheless, it's evident that current SOTA image models still struggle with consistent positional understanding.

Overall Flux.dev performed significantly better than DALLE-3, demonstrating an average misalignment rate of 59\% whilst DALLE-3 performed significantly worse at 71\%. There were some standout differences in some categories, namely Associative(AND), Commutative(Y), and DeMorgan(Y). DALLE-3 was observed to prefer generating multiple of an entity when only prompted for one, as well as significant misunderstanding in vertical positional prompts.

\vspace{-2mm}
\subsection{Common misalignment errors}
\begin{figure}[t]
    \centering
    \includegraphics[scale=0.35]{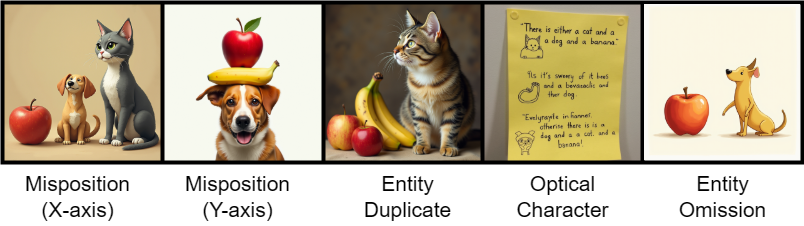}
    \vspace{-3mm}
    \caption{Examples of misalignment errors categorised under 5 labels: X-Misposition, Y-Misposition, Entity Duplication, Optical Character and Entity Omission.} 
    \label{fig:errors}
    \vspace{-3mm}
\end{figure}

\vspace{-2mm}
By performing a manual inspection of the saved misalignment image pairs, we are able to identify distinct categories of misalignments detected by MetaLogic. We classified these errors into five main types: (1) X-Axis misposition, (2) Y-axis misposition, (3) Entity duplication, (4) Optical character, and (5) Entity omission (See Figure \ref{fig:errors}). This classification provides insight into the specific challenges text-to-image models face when interpreting logical relationships.

Both entity duplication and entity omission were identified in almost all logic categories and variations. This is likely a result of failure in logic reasoning that manifests as failure to generate a specific object or mistakenly doubling up on a singular entity.

As expected, the spatial mispositioning of entities across both horizontal and vertical axis appeared only in image pairs generated from X and Y variants of prompt pairs. This result was already anticipated as those were the only prompts that incorporated positional information within its semantic equivalence. However, additional information within the prompts had little to no impact on the average misalignment rate compared to its baseline. We believe a proportion of the images seeding at inference are weaker in understanding both conjunctive and positional semantics, resulting in a similar failure rate for both categories.

The most unexpected error type we observed was Optical Character Generation, where the model resorted to generating text explanations within the image rather than properly representing the logical relationship visually. This occurred predominantly with long and complex prompts such as Associative (OR) and Distributive (OR), which contained embedded Commutative logic to maintain a singular semantic equivalence. Both models consistently struggled to understand the underlying logic within the prompt and fell back on simply generating parts, or all of the initial prompt within the image. As shown in Figure \ref{fig:errors}, when faced with these complex logical structures, the model occasionally generated explanatory text or diagrams within the image, effectively "giving up" on pure visual representation in favour of textual explanation. This suggests a fallback mechanism when the model's visual generation capabilities are unable to handle complex logical operations.

\vspace{-3mm}
\subsection{Numbering entities}
\begin{figure}
    \centering
    \includegraphics[width=0.75\linewidth]{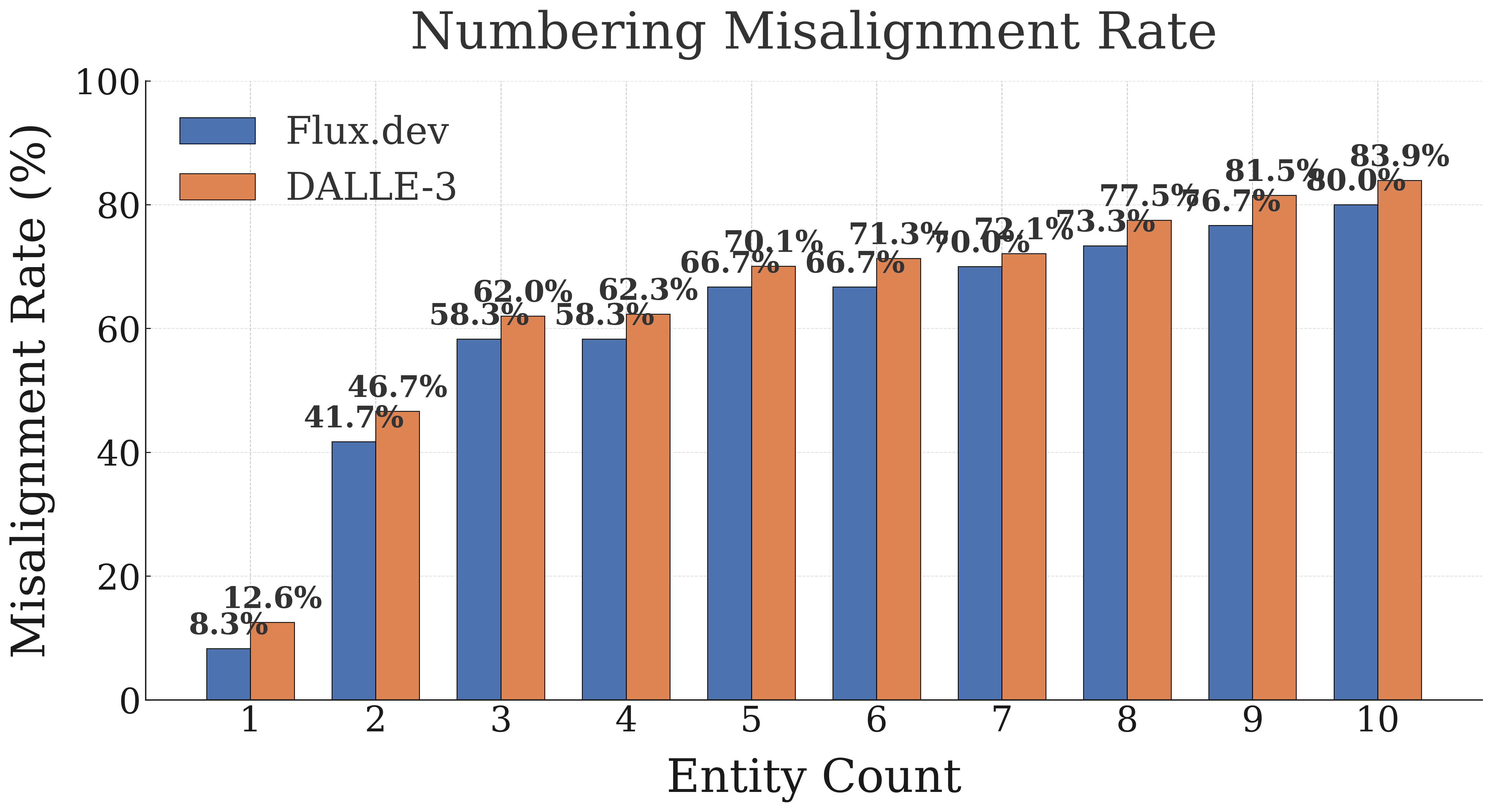}
    \vspace{-4mm}
    \caption{The misalignment rate for varying entity counts from 1 to 10 of (cat, dog, apple, banana) using Flux.dev and DALLE-3.}
    \label{fig:number_entities}
    \vspace{-6mm}
\end{figure}

\vspace{-2mm}
Figure \ref{fig:number_entities} illustrates a clear trend, as the number of entities in a prompt increases, misalignment rates increase across both models. Prompts with a single entity maintain low misalignment (10\%), but performance declines steeply beyond two entities. By the time prompts include seven or more entities, alignment often drops below 30\% on average.

It was also observed in misalignment output directory that the trajectory of this decline varies by entity type. While “cat,” “dog,” and “apple” show relatively stable degradation, “banana” exhibits a significantly sharper increase in misalignment, even at lower counts. This may stem from its typical representation in training data, where bananas appear in bunches rather than individually. Such visual-semantic biases likely interfere with logical interpretation, underscoring that model errors arise not only from prompt complexity but also from mismatches between learned visual associations and logical intent.

\vspace{-3mm}
\section{Discussion}
\vspace{-2mm}
Our findings reveal fundamental gaps in how current text-to-image models process logical relationships. The consistent pattern of increasing misalignment rates as logical complexity increases from simple commutative to complex distributive relationships, suggests that these models lack robust logical reasoning capabilities despite their impressive visual generation abilities.

The high failure rates with OR statements (75\% misalignment versus 46\% for AND statements) indicate that disjunctive logic presents a particular challenge. This may stem from the inherent ambiguity in training data, where multiple valid visual interpretations exist for prompts containing ``or'' relationships. Model architectures may benefit from explicit training on logical equivalence relationships, potentially through contrastive learning approaches that minimise differences between outputs from logically equivalent prompts.
The unexpected emergence of text generation within images when faced with complex logic (observed in 8\% of cases) represents a fascinating fallback mechanism. This suggests that when the model's visual generation capabilities cannot adequately represent the logical relationship, it defaults to its text generation capabilities, essentially ``explaining'' rather than ``showing'' the relationship. This insight could inform hybrid approaches that leverage both visual and textual representation for complex prompts.

The identification of logical reasoning gaps in text-to-image models has implications beyond technical improvement. As these models increasingly serve as interfaces between human intent and visual creation, logical consistency becomes crucial for applications in education, design, and communication. Misalignments in logical understanding could lead to miscommunication or even potentially harmful outputs in sensitive contexts.
By providing a method to systematically identify and categorise these misalignments, MetaLogic contributes to the broader goal of making generative AI systems more reliable, predictable, and aligned with human intent. The counterexamples generated through our approach not only highlight current limitations but also provide concrete materials for improving model transparency and user understanding of these systems' capabilities and limitations.

\vspace{-4mm}
\section{Limitations}
\vspace{-2mm}
The core innovation of MetaLogic, eliminating the need for external ground truth through semantically equivalent prompt pairs inherently creates specific limitations in evaluation scope.
While our approach successfully identifies misalignment without requiring a predetermined ground truth, it fundamentally assumes that logically equivalent prompts should produce semantically identical images. This assumption breaks down in two key scenarios:
\begin{itemize}
    \item First, prompts containing disjunctive statements (OR) pose a theoretical challenge, since multiple distinct visual outputs can all be valid. For example, “There is a cat or a dog” could yield either only a cat or only a dog, making differences between logically equivalent prompts ambiguous rather than reveal true misalignments. To address this, we introduced workarounds using embedded Commutative equivalence, though symptoms of this solution are apparent in the misalignment rates found in the Distributive category where these unwarranted complexities potentially skewed the final results as theoretically, AND and OR prompts require similar levels of logical reasoning to understand.

    \item Second, our approach cannot evaluate true negation. Statements like "There is no cat" have theoretically infinite correct visual representations, as any scene without a cat would be valid. Since there is no single expected visual output for negation statements, the semantic equivalence comparison mechanism breaks down entirely. The assumption that semantic equivalence in prompts should produce visual consistency in outputs does not hold when the space of valid outputs is unbounded. Our approach circumvents this by applying a double negative to return the prompt back to a singular correct output, but still falls short of true negation statements.

    \item Lastly, our framework remains dependent on strong object-detection capabilities. Whilst Florence-2 demonstrates strong performances from our manual checks, occasional inaccuracies can still influence results. This reliance highlights the need for complementary validation strategies or future benchmarks that can assess detection accuracy on generated images.
\end{itemize}

These limitations highlight an important trade-off in our reference-free approach: by eliminating the need for external ground truth, we necessarily constrain our evaluation scope to logical relationships that produce well-defined, consistent visual expectations across equivalent prompts. Future work will explore hybrid approaches that maintain MetaLogic's efficiency while addressing these edge cases.

\vspace{-4mm}
\section{Conclusion}
\vspace{-2mm}
In this paper, we present MetaLogic, a framework for evaluating text-to-image model robustness via metamorphic testing with logically equivalent prompts. Unlike conventional reference-based methods, MetaLogic compares image pairs from grammatically different but semantically equivalent prompts, enabling scalable, ground truth-free robustness assessment.
Experiments on state-of-the-art models (Flux.dev, DALLE-3) reveal significant robustness failures under minor logic-preserving variations. Misalignment rates increase with logical complexity: Distributive (86\%), Associative (73\%), DeMorgan (59\%), Complement (52\%), and Commutative (54\%), with a stark contrast between conjunctive and disjunctive forms (45\% vs. 62\%). Common errors include entity omission, duplication, positional shifts, and unexpected text generation. These counterexamples offer actionable insights for model improvement with lower computational cost than existing evaluations. As T2I models are adopted in real-world applications, MetaLogic provides a principled and efficient robustness evaluation, advancing logically consistent and trustworthy generative systems.

\bibliography{mybibliography}   

\begin{thebibliography}{10}
\providecommand{\url}[1]{\texttt{#1}}
\providecommand{\urlprefix}{URL }
\providecommand{\doi}[1]{https://doi.org/#1}

\bibitem{ahmadi2023examination}
Ahmadi, S., Agrawal, A.: An examination of the robustness of reference-free image captioning evaluation metrics. arXiv preprint arXiv:2305.14998  (2023)

\bibitem{betker2023improving}
Betker, J., Goh, G., Jing, L., Brooks, T., Wang, J., Li, L., Ouyang, L., Zhuang, J., Lee, J., Guo, Y., et~al.: Improving image generation with better captions  \textbf{2}(3), ~8 (2023), \url{https://cdn.openai.com/papers/dall-e-3.pdf}

\bibitem{bronkhorst2020logical}
Bronkhorst, H., Roorda, G., Suhre, C., Goedhart, M.: Logical reasoning in formal and everyday reasoning tasks. International Journal of Science and Mathematics Education  \textbf{18},  1673--1694 (2020), \url{https://doi.org/10.1007/s10763-019-10039-8}

\bibitem{chen2020metamorphictestingnewapproach}
Chen, T.Y., Cheung, S.C., Yiu, S.M.: Metamorphic testing: A new approach for generating next test cases (2020), \url{https://arxiv.org/abs/2002.12543}

\bibitem{10.1145/3143561}
Chen, T.Y., Kuo, F.C., Liu, H., Poon, P.L., Towey, D., Tse, T.H., Zhou, Z.Q.: Metamorphic testing: A review of challenges and opportunities. ACM Comput. Surv.  \textbf{51}(1) (Jan 2018). \doi{10.1145/3143561}, \url{https://doi.org/10.1145/3143561}

\bibitem{conwell2024relations}
Conwell, C., Tawiah-Quashie, R., Ullman, T.: Relations, negations, and numbers: Looking for logic in generative text-to-image models. arXiv preprint arXiv:2411.17066  (2024)

\bibitem{fu2024commonsense}
Fu, X., He, M., Lu, Y., Wang, W.Y., Roth, D.: Commonsense-t2i challenge: Can text-to-image generation models understand commonsense? arXiv preprint arXiv:2406.07546  (2024)

\bibitem{ghosh2024geneval}
Ghosh, D., Hajishirzi, H., Schmidt, L.: Geneval: An object-focused framework for evaluating text-to-image alignment. Advances in Neural Information Processing Systems  \textbf{36} (2024)

\bibitem{grimal2024tiam}
Grimal, P., Le~Borgne, H., Ferret, O., Tourille, J.: Tiam-a metric for evaluating alignment in text-to-image generation. In: Proceedings of the IEEE/CVF Winter Conference on Applications of Computer Vision. pp. 2890--2899 (2024). \doi{10.1109/wacv57701.2024.00287}

\bibitem{hu2023tifa}
Hu, Y., Liu, B., Kasai, J., Wang, Y., Ostendorf, M., Krishna, R., Smith, N.A.: Tifa: Accurate and interpretable text-to-image faithfulness evaluation with question answering. In: Proceedings of the IEEE/CVF International Conference on Computer Vision. pp. 20406--20417 (2023). \doi{10.1109/iccv51070.2023.01866}

\bibitem{huang2022towards}
Huang, J., Chang, K.C.C.: Towards reasoning in large language models: A survey. arXiv preprint arXiv:2212.10403  (2022)

\bibitem{flux2024}
Labs, B.F.: Flux. \url{https://github.com/black-forest-labs/flux} (2024)

\bibitem{lin2014microsoft}
Lin, T.Y., Maire, M., Belongie, S., Hays, J., Perona, P., Ramanan, D., Doll{\'a}r, P., Zitnick, C.L.: Microsoft coco: Common objects in context. In: Computer Vision--ECCV 2014: 13th European Conference, Zurich, Switzerland, September 6-12, 2014, Proceedings, Part V 13. pp. 740--755. Springer (2014), \url{https://doi.org/10.1007/978-3-319-10602-1}

\bibitem{mirzadeh2024gsm}
Mirzadeh, I., Alizadeh, K., Shahrokhi, H., Tuzel, O., Bengio, S., Farajtabar, M.: Gsm-symbolic: Understanding the limitations of mathematical reasoning in large language models. arXiv preprint arXiv:2410.05229  (2024)

\bibitem{Mo_2024_CVPR}
Mo, W., Zhang, T., Bai, Y., Su, B., Wen, J.R., Yang, Q.: Dynamic prompt optimizing for text-to-image generation. In: Proceedings of the IEEE/CVF Conference on Computer Vision and Pattern Recognition (CVPR). pp. 26627--26636 (June 2024), \url{http://dx.doi.org/10.1109/CVPR52733.2024.02514}

\bibitem{Nichol2021GLIDETP}
Nichol, A., Dhariwal, P., Ramesh, A., Shyam, P., Mishkin, P., McGrew, B., Sutskever, I., Chen, M.: Glide: Towards photorealistic image generation and editing with text-guided diffusion models. In: International Conference on Machine Learning (2021), \url{https://api.semanticscholar.org/CorpusID:245335086}

\bibitem{radford2021learning}
Radford, A., Kim, J.W., Hallacy, C., Ramesh, A., Goh, G., Agarwal, S., Sastry, G., Askell, A., Mishkin, P., Clark, J., et~al.: Learning transferable visual models from natural language supervision. In: International conference on machine learning. pp. 8748--8763. PMLR (2021), \url{https://arxiv.org/abs/2103.00020}

\bibitem{ramesh2022hierarchical}
Ramesh, A., Dhariwal, P., Nichol, A., Chu, C., Chen, M.: Hierarchical text-conditional image generation with clip latents. arXiv preprint arXiv:2204.06125  \textbf{1}(2), ~3 (2022)

\bibitem{Saharia2022PhotorealisticTD}
Saharia, C., Chan, W., Saxena, S., Li, L., Whang, J., Denton, E.L., Ghasemipour, S.K.S., Ayan, B.K., Mahdavi, S.S., Lopes, R.G., Salimans, T., Ho, J., Fleet, D.J., Norouzi, M.: Photorealistic text-to-image diffusion models with deep language understanding. ArXiv  \textbf{abs/2205.11487} (2022), \url{https://api.semanticscholar.org/CorpusID:248986576}

\bibitem{8573811}
Segura, S., Towey, D., Zhou, Z.Q., Chen, T.Y.: Metamorphic testing: Testing the untestable. IEEE Software  \textbf{37}(3),  46--53 (2020). \doi{10.1109/MS.2018.2875968}

\bibitem{singh2023divide}
Singh, J., Zheng, L.: Divide, evaluate, and refine: Evaluating and improving text-to-image alignment with iterative vqa feedback. Advances in Neural Information Processing Systems  \textbf{36},  70799--70811 (2023)

\bibitem{sueyoshi2024predicated}
Sueyoshi, K., Matsubara, T.: Predicated diffusion: Predicate logic-based attention guidance for text-to-image diffusion models. In: Proceedings of the IEEE/CVF Conference on Computer Vision and Pattern Recognition. pp. 8651--8660 (2024), \url{https://doi.org/10.1109/cvpr52733.2024.00826}

\bibitem{wan-etal-2024-logicasker}
Wan, Y., Wang, W., Yang, Y., Yuan, Y., Huang, J.T., He, P., Jiao, W., Lyu, M.: Logicasker: Evaluating and improving the logical reasoning ability of large language models. pp. 2124--2155 (01 2024). \doi{10.18653/v1/2024.emnlp-main.128}

\bibitem{xiao2024florence}
Xiao, B., Wu, H., Xu, W., Dai, X., Hu, H., Lu, Y., Zeng, M., Liu, C., Yuan, L.: Florence-2: Advancing a unified representation for a variety of vision tasks. In: Proceedings of the IEEE/CVF Conference on Computer Vision and Pattern Recognition. pp. 4818--4829 (2024), \url{https://doi.org/10.1109/cvpr52733.2024.00461}

\bibitem{10531644}
Xu, L., Bu, X., Tian, X.: Dynamic prompt-driven zero-shot relation extraction. IEEE/ACM Transactions on Audio, Speech, and Language Processing  \textbf{32},  2900--2912 (2024). \doi{10.1109/TASLP.2024.3402063}

\bibitem{yang2023dynamic}
Yang, F., Yang, S., Butt, M.A., van~de Weijer, J., et~al.: Dynamic prompt learning: Addressing cross-attention leakage for text-based image editing. Advances in Neural Information Processing Systems  \textbf{36},  26291--26303 (2023), \url{https://doi.org/10.48550/arXiv.2309.15664}

\bibitem{Yuan_2021_CVPR}
Yuan, Y., Wang, S., Jiang, M., Chen, T.Y.: Perception matters: Detecting perception failures of vqa models using metamorphic testing. In: Proceedings of the IEEE/CVF Conference on Computer Vision and Pattern Recognition (CVPR). pp. 16908--16917 (June 2021), \url{https://doi.org/10.1109/cvpr46437.2021.01663}

\end{thebibliography}
\bibliographystyle{splncs04}

\end{document}